\pdfoutput=1

\documentclass[11pt]{article}
\usepackage[usenames,dvipsnames]{color}
\usepackage{acl}

\usepackage{times}
\usepackage{latexsym}

\usepackage[T1]{fontenc}

\usepackage[utf8]{inputenc}

\usepackage{tikz}
\usepackage{amsmath}
\usepackage{subfig}
\usepackage{tcolorbox}

\usepackage{graphicx}
\usepackage{multirow}
\usepackage{booktabs}
\usepackage{amsfonts}
\usepackage{verbatim}
\usepackage{pgfplots}
\usepackage{wrapfig}
\usepackage{amssymb,mathrsfs}
\usepackage{amssymb}
\usepackage{array}
\usepackage{pgfplotstable}
\usepackage{pgf}
\usepackage{bm}
\usepackage[normalem]{ulem}
\usepackage{colortbl}
\usepackage{soul}
\usepackage{ulem}
\usepackage{xspace}

\usetikzlibrary{plotmarks}
\usetikzlibrary{backgrounds}
\usetikzlibrary{fit}
\usetikzlibrary{decorations}
\usetikzlibrary{decorations.markings}
\usetikzlibrary{positioning}

\definecolor{myblue}{RGB}{31,120,180}
\definecolor{mygreen}{RGB}{51,160,44}
\definecolor{myred}{RGB}{227,26,28}

\definecolor{iyellow}{RGB}{255,250,205}
\definecolor{ipurple}{RGB}{230,230,250}

\definecolor{upurple}{RGB}{155,89,182}
\definecolor{ublue}{RGB}{52,152,219}
\definecolor{ured}{RGB}{231,76,60}
\definecolor{udark}{RGB}{77,153,77}
\definecolor{ugreen}{RGB}{46,204,113}

\usepackage{microtype}

\newcommand{\MR}[3]{\multirow{#1}{#2}{#3}}

\newcommand{\B}{\textbf}
\newcommand{\I}{\textit}
\newcommand{\T}{\texttt}

\title{On Vision Features in Multimodal Machine Translation}

\author{
  Bei Li$^1$,
  Chuanhao Lv$^1$,
  Zefan Zhou$^1$,
  Tao Zhou$^1$,\\
  \textbf{Tong Xiao$^{1,2}$\thanks{\xspace\xspace Corresponding author.}},
  \textbf{Anxiang Ma$^{1,2}$}
  \textbf{and Jingbo Zhu$^{1,2}$}\\
  $^{1}$School of Computer Science and Engineering, Northeastern University, Shenyang, China\\
  $^{2}$NiuTrans Research, Shenyang, China \\
  {\tt
        \{libei\_neu,lch-sy,ZhouZefan\_zzf,zhoutao\_neu\}@outlook.com
  }\\
  {\tt
		\{xiaotong,maanxiang,zhujingbo\}@mail.neu.edu.cn
  }
}

\begin{document}
\maketitle
\begin{abstract}
	Previous work on multimodal machine translation (MMT) has focused on the way of incorporating vision features into translation but little attention is on the quality of vision models. In this work, we investigate the impact of vision models on MMT. Given the fact that Transformer is becoming popular in computer vision, we experiment with various strong models (such as Vision Transformer) and enhanced features (such as object-detection and image captioning). We develop a selective attention model to study the patch-level contribution of an image in MMT. On detailed probing tasks, we find that stronger vision models are helpful for learning translation from the visual modality. Our results also suggest the need of carefully examining MMT models, especially when current benchmarks are small-scale and biased. Our code could be found at \url{https://github.com/libeineu/fairseq_mmt}.
	\end{abstract}
	
	\section{Introduction}
	
	Multimodal machine translation (MMT) has emerged as an active field of research which marries the worlds of computer vision (CV) and natural language processing (NLP) \cite{specia-etal-2016-shared}. Early models of this kind produce a translation given the fused representation of both the visual and textual inputs \cite{caglayan2016multimodal,libovicky-helcl-2017-attention,calixto-liu-2017-incorporating}. As expected, such a paradigm achieves promising BLEU improvements and inspires the community to follow up.
	
	But soon researchers found that MMT systems did not act as what they ordinarily designed: the visual modality contributes to translation little. For example, it is not harmful to MMT systems when the input image is irrelevant to the text \cite{gronroos-etal-2018-memad,lala-etal-2018-sheffield}, or even when the vision features are absent \cite{elliott-2018-adversarial}. More recently, \citet{wu-etal-2021-good} have pointed out that the use of the visual modality is a way of regularization for training but not a complement to the text modality. As another response to the analysis of MMT, \citet{caglayan-etal-2019-probing} investigate how the vision features correlate to the text. They find that the input image helps translation when some of the input words are masked.
	
	Note that previous work has for the most part focused on integrating off-the-shelf vision models (such as ResNet-50) into MMT. The underlying assumption here is that the existing vision models are powerful enough to encode the image. This implicitly ignores the quality of vision models in representing images. But computer vision is facing a new trend by moving from CNNs to Transformer as the backbone model \cite{dosovitskiy2021ViT,liu2021swin,carion2020DETR}. A natural question that arises is: \textit{how will MMT systems behave if stronger vision models are adopted?}
	
	In this work, we address this question by a systematic study of using various vision models in MMT, in particular using the most successful models in recent studies (such as Vision Transformer, or ViT for short). We find that the patch method used in Transformer-based vision models offers an opportunity to detail the patch-level contribution of the image. This leads us to develop a selective attention model to correlate words with image patches. Beyond this, we introduce object-detection and image captioning features into MMT for further improvements of the vision models \cite{carion2020DETR,fang2021QueryInst}.

\begin{table*}
	\begin{minipage}[!t]{.1\linewidth}
	  \centering
	  \includegraphics[width=2.2cm,height=2.9cm]{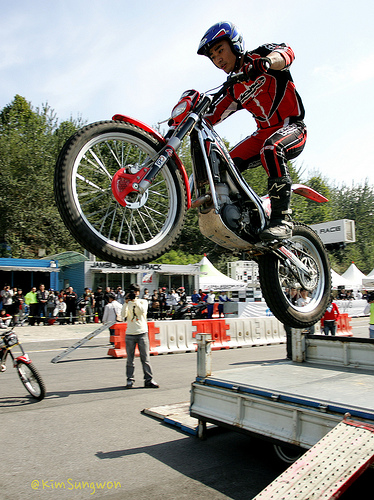}
	\end{minipage}
	\begin{minipage}[h]{.9\linewidth}
		\centering
		\small
		\setlength\tabcolsep{4.2pt}
		\begin{tabular}{lcccccccccc}
		
		\toprule
		\T{SRC} :& \bf a & \bf man& \bf in& \bf a & \bf red & \bf suit &\bf performing & \bf motorcycle & \bf stunts \\
		\midrule
		\T{Color} &a & man & in& a & [$\texttt{MASK\_C}$] &suit & performing & motorcycle &  stunts\\
		\T{Char.} &a & [$\texttt{MASK\_P}$] & in& a & red &suit & performing & motorcycle &  stunts  \\
		\T{MASK1} &a & man & in& a & red &[$\texttt{MASK\_N}$] & performing & motorcycle &  stunts \\
		\T{MASK2} &a & man & in& a & red &[$\texttt{MASK\_N}$] & performing & [$\texttt{MASK\_N}$] &  stunts  \\
		\T{MASK3} &a & man & in& a & red &[$\texttt{MASK\_N}$] & performing & [$\texttt{MASK\_N}$] &  [$\texttt{MASK\_NS}$] \\
		\T{MASK4} &a & [$\texttt{MASK\_N}$] & in& a & red &[$\texttt{MASK\_N}$] & performing & [$\texttt{MASK\_N}$] &  [$\texttt{MASK\_NS}$]\\
	
		\bottomrule
		\end{tabular}
	\end{minipage}
	 \caption{An example of the proposed probing tasks. We replace the masked token by four symbols respectively.}
	\label{tab:insufficient_text}
\end{table*}

		

Following \citet{caglayan-etal-2019-probing}'s work, we design more detailed probing tasks to examine to what degree the visual modality contributes to MMT. We run an extensive set of experiments on En-De and En-Fr MMT tasks. Our findings are

\begin{itemize}
\item Stronger vision models help. For example, ViT can beat ResNet-50 on the probing tasks though the superiority is not significant on standard MMT data.
\item Automatic evaluation on current MMT tasks might not be a good indicator for the effectiveness of MMT models. For example, models enhanced with object-detection and image captioning features yield good BLEU scores on the original MMT task but show modest or no contributions on the probing tasks.
\end{itemize}

We hope that the results here can inspire more research on exploring better vision models and evaluation methods for multimodal NLP.

\section{Preliminary}

We start with a description of the probing tasks. It is followed by a design of vision features and a selective attention mechanism for introducing ViT-like representations into MMT.

\subsection{Insufficient Text Generation}
\label{sec:masking}

To know how an image contributes to translation, a way is to mask some of the input words (call this insufficient text) and force the translation model to learn from the image. Following the previous design of color deprivation and entity-based masking, we present detailed probing tasks which are complementary to \citet{caglayan-etal-2019-probing}'s work. In preliminary experiments\footnote{We choose the Multi30K En-De and En-Fr datasets for experiments.}, we find that ``color'', ``character'' and ``noun'' are three kinds of words which could be complemented according to the visual modality once the corresponding texts are masked. The following probing tasks are designed accordingly.

\paragraph{Color-based Probing} In training, all source words referring to a color are replaced by a special token $[\texttt{Mask\_C}]$. There are $8,919$ sentences involving color words, and nearly one third of them involve more than one color. It is worth noting that each color may have two or more translations due to the rich morphology in German and French. For example, the English ``green'' can be translated to ``grün'', ``grüne'', ``grünes'', ``grüner'', ``grünen'' and ``grünem'' in German. We design two criteria to measure the accuracy of translation. The first criterion is strict. The correct translation requires generating the same color and the same gender as in reference translations. The second criterion is relaxed and all translations expressing the same color are correct.

\paragraph{Character-based Probing}
For character words, we choose ``man'', ``woman'', ``people'', ``men'', ``girl'' and ``boy''. More than $60\%$ sentences contain character words in our training data, so they are reasonable indicators of assessing the ability to infer correct translations from the input image. Here we use [$\texttt{MASK\_P}$] for masking. Note that some character words have more than two translations, e.g. ``people'', we also use the same evaluation metric with the color-based probing task, including relaxed and strict two criteria.

\paragraph{Noun-based Probing}
For more complex scenarios, a sentence can be masked with several kinds of ambiguous words, such as animals, clothing, and vehicles, provided by Flickr30K \cite{plummer2015flickr30k}. High-frequency words labeled with noun (or nouns) are more likely to be masked as [$\texttt{MASK\_N}$] (or [$\texttt{MASK\_NS}$])). See Table \ref{tab:insufficient_text} for example insufficient text with different numbers of masks.

\subsection{Various Vision Features}
In addition to ResNet-50, we choose several Transformer-based vision models.
\input{fig_selective_attn}
\begin{itemize}
	\item General Backbone. Vision Transformer (ViT) and Swin Transformer are popular models in computer vision \cite{dosovitskiy2021ViT,liu2021swin}. We use ViT with various model capacities to vary from weak to strong ViT models.
	\item Object-detection. For pretrained object-detection vision models, we choose DETR \cite{carion2020DETR} and QueryInst \cite{fang2021QueryInst} for their strong performance.
	\item Image Captioning. For image captioning models, we choose CATR\footnote{\url{https://github.com/saahiluppal/catr}} because it is a Transformer-based image captioning architecture and can be easily implemented on top of ViT.
\end{itemize}
	
We form a number of vision features by combining the methods described above. More details are presented in Section \ref{sec:experiments}.

\subsection{Selective Attention}

ViT and related models perform in almost the same way as Transformer in NLP \cite{vaswani2017attention}. Unlike the general models in CV, ViT does not represent the image as a single vector. Instead, it generates a sequence of patches for image representation. An advantage of this design is that we can use the attention mechanism to correlate image patches to words. Thus, we present a selective attention model to model the patch-level contribution of the image. See Figure \ref{fig:selective_attn} for the architecture.

\paragraph{Text-only Transformer}

Transformer follows an encoder-decoder paradigm (the purple region in Figure \ref{fig:selective_attn}) . The encoder is a stack of identical layers. Each layer consists of a self-attention (SAN) block and a feedforward network (FFN) block. The decoder shares a similar design with the encoder, but with an additional cross-attention block.

\begin{table*}[]
	\centering
	\resizebox{\textwidth}{!}{
		\begin{tabular}{c|l|l|c|c|c|c|c|c|c|c|c|c|c|c}
		\hline
		\multirow{2}{*}{\#} &\multirow{2}{*}{\textbf{Model}} &\multirow{2}{*}{\textbf{Feature}} & \multicolumn{6}{c|}{\textbf{English$\to$German}} & \multicolumn{6}{c}{\textbf{English$\to$French}} \\
		\cline{4-15}
		& & &\multicolumn{2}{c|}{\textbf{Test2016}} & \multicolumn{2}{c|}{\textbf{Test2017}} & \multicolumn{2}{c|}{\textbf{MSCOCO}} &\multicolumn{2}{c|}{\textbf{Test2016}} & \multicolumn{2}{c|}{\textbf{Test2017}} & \multicolumn{2}{c}{\textbf{MSCOCO}} \\
		\hline
		\multicolumn{15}{c}{\textit{Text-only Transformer}} \\ \hline
		1 &\textbf{Tiny}  & - & 41.02 & 68.22 & 33.36 & 62.05 & 29.88 & 56.64 & 61.80  & 81.02 & 53.46 & 75.62 & 44.52 & 69.43 \\
		\hline
		\multicolumn{15}{c}{\textit{Existing MMT Systems}} \\
		\hline
		2 &\textbf{Doubly-ATT}        & ResNet & 41.45 & 68.04 & 33.95 & 61.83 & 29.63 & 56.21 & 61.99  & 81.12 & 53.72 & 75.71 & 45.16 & 70.25 \\
	    3 &\textbf{Imagination}       & ResNet & 41.31 & 68.06 & 32.89 & 61.29 & 29.90 & 56.57 & 61.90  & 81.20 & 54.07 & 76.03 & 44.81 & 70.35 \\
		4 &\textbf{UVR-NMT}           & ResNet & 40.79 & -     & 32.16 & -     & 29.02 & -     & 61.00  & -     & 53.20 & -     & 43.71 & -     \\
		5 &\textbf{Gated Fusion}      & ResNet & 41.96 & 67.84 & 33.59 & 61.94 & 29.04 & 56.15 & 61.69  & 80.97 & 54.85 & 76.34 & 44.86 & 70.51 \\
		\hline
		\multicolumn{15}{c}{\textit{Our MMT Systems}} \\
		\hline
		6 &\textbf{Gated Fusion}   & ViT-Large    & 41.55 & 68.34 & 33.49 & 61.67 & 29.27 & 55.64 & 61.93 & 81.08 & 54.98 & 75.12 & 45.65 & 70.81 \\
		7 &\textbf{Selective Attn} & ViT-Large    & 41.84 & 68.64 & 34.32 & 62.32 & 30.22 & 56.91 & 62.24 & 81.41 & 54.52 & 76.30 & 44.82 & 70.63    \\
		\hline
		8 &\textbf{7 + ViT-Tiny}	& ViT-Tiny  & 40.74 & 67.20 & 32.48 & 60.46 & 28.10 & 55.19 & 61.44  & 80.91 & 53.31 & 75.65 & 45.82 & 70.75 \\
		9 &\textbf{7 + ViT-Small}	& ViT-Small & 40.86 & 67.64 & 33.62 & 61.61 & 29.72 & 56.94 & 61.78  & 81.30 & 54.21 & 76.04 & 45.28 & 70.89 \\
		10 &\textbf{7 + ViT-Base}	& ViT-Base  & 41.93 & 68.55 & 33.60 & 61.42 & 31.14 & 56.77 & 62.48  & 81.71 & 54.44 & 76.46 & 44.72 & 71.20 \\
		\hline
		11 &\textbf{7 + DETR}		& DETR  		& 42.23 & 68.94 & 34.14 & 61.57 & 30.13 & 57.01 & 62.14  & 81.45 & 55.17 & 76.40 & 45.10 & 70.38 \\	
		12 &\textbf{7 + QueryInst}	& QueryInst  	& 41.90 & 68.64 & 34.90 & 62.27 & 30.20 & 56.89 & 62.33  & 81.26 & 54.97 & 76.61 & 45.56 & 70.64 \\
		13 &\textbf{7 + CATR}		& CATR  		& 42.50 & 68.81 & 34.28 & 61.81 & 29.59 & 56.36 & 62.79  & 81.75 & 55.44 & 76.57 & 45.27 & 70.73 \\
		\hline
		\end{tabular}
		}
		\caption{BLEU (left) and METEOR (right) scores of En$\to$De and En$\to$Fr tasks. Some of the results are from \citet{wu-etal-2021-good}'s work.}
		\label{tab:multi30k}
	\end{table*}

\paragraph{Gated Fusion}
Gated fusion mechanism is a popular technique for fusing representations from different sources \cite{wu-etal-2021-good,zhang2020UVR,lin2020capsule,yin-etal-2020-novel}. Given the text input $X^{\textrm{text}}$ and the image input $X^{\textrm{img}}$, the text representation $H^{\textrm{text}}$ and the image feature $H^{\textrm{img}}$ can be defined as:

\begin{eqnarray}
	H^{\textrm{text}} & = & \textrm{TransformerEncoder}(X^{\textrm{text}}) \\
	H^{\textrm{img}}  & = & W \  \textrm{ViT}(X^{\textrm{img}})
	\label{eq:projection}
\end{eqnarray}

\noindent where $W$ is a projection matrix to convert the shape of $\textrm{ViT}(X^{\textrm{img}})$ into that of $H^{\textrm{text}}$. Note that $\textrm{ViT}(\cdot)$ can be replaced by other vision models, e.g. DETR, Swin Transformer and etc. Then,
the gate $\lambda\in[0,1]$ and the fuzed output are defined as:
\begin{eqnarray}
	\lambda & = & \textrm{Sigmoid}(U{H^{\textrm{text}}} + V{H^{\textrm{img}}}) \label{eq:gate}\\
	H^{\textrm{Out}} & = & (1 - \lambda) \cdot H^{\textrm{text}} + \lambda \cdot H^{\textrm{img}} \label{eq:gated_fusion}
\end{eqnarray}
\noindent where $U$ and $V$ are trainable variables. $\lambda$ controls how much visual information is kept. Then, the fusion vector $H^{\textrm{Out}}$ is fed into the decoder. See the right side of the pink region in Figure \ref{fig:selective_attn} for an illustration of the gated fusion models.

\paragraph{Selective Attention}

After obtaining the text and image representations (or features), we use a single-head attention network to correlate words with image patches, where the query, key and value are $H^{\textrm{text}}$, $H^{\textrm{img}}$ and $H^{\textrm{img}}$, respectively. Then the selective attention output $H^{\textrm{img}}_{attn}$ is defined to be:

\begin{eqnarray}
	H^{\textrm{img}}_{attn} & = & \textrm{Softmax}(\frac{Q{K^{\textrm{T}}}}{\sqrt{d_k}})V
	  \label{eq:selective_attn}
\end{eqnarray}

\noindent where $d_k$ is the same as the dimension of $H^{\textrm{text}}$ because a single head is used. Then the fused representation could be obtained by using Eqs. \ref{eq:gate} and \ref{eq:gated_fusion}  and replacing $H^{\textrm{img}}$ with $H^{\textrm{img}}_{attn}$.

\section{Experiments}
\label{sec:experiments}

\subsection{Datasets}
We conducted experiments on the widely used Multi30K benchmark \cite{elliott2016Multi30k}. The training and validation sets consisted of $29,000$ and $1,014$ instances, respectively. We reported the results on the Test2016, Test2017 and MSCOCO test sets \cite{elliott-etal-2017-findings}. Note that MSCOCO is more challenging for MMT models due to the out-of-domain instances with ambiguous verbs. Following the setup in \cite{wu-etal-2021-good}, we learned a joint BPE code for $10,000$ merging operations for both the source and target languages, resulting in vocabularies of $9,716$ and $9,548$ entries for the En-De and En-Fr tasks.

\subsection{Experimental Setups}
We followed the \citet{wu-etal-2021-good}'s work to conduct experiments with Transformer-Tiny configuration, which is more suited for small datasets like Multi30K. Note that smaller models even obtain higher BLEU scores than pervious MMT models. Similar observations have been discussed when building context-aware machine translation models \cite{li-etal-2020-multi-encoder}. The model consists of $4$ encoder and decoder layers. The hidden size is $128$ and the filter size of FFN is $256$. There are $4$ heads in the multi-head self-attention mechanism. We set the dropout as $0.3$ and the label smoothing as $0.1$.

\begin{table*}[t]
	\resizebox*{\linewidth}{!}{
	\centering
	\begin{tabular}{l|ll|ll|ll}
	\hline
	\multirow{2}{*}{\textbf{Systems}} & \multicolumn{2}{c|}{\textbf{Test2016}} & \multicolumn{2}{c|}{\textbf{Test2017}} & \multicolumn{2}{c}{\textbf{MSCOCO}} \\
	\cline{2-7}
	& \textbf{Restrict} & \textbf{Relaxed} & \textbf{Restrict} & \textbf{Relaxed} & \textbf{Restrict} & \textbf{Relaxed}\\
	\hline
	\multicolumn{7}{c}{\textit{English$\to$German}} \\
	\hline
	Text-only Transformer  &25.93                   &34.42                    &22.57                   &35.70                    &18.75    				  &23.44\\
	Gated Fusion + ResNet  &27.23 ($\uparrow$ 1.30) &35.51 ($\uparrow$ 1.09)  &23.10 ($\uparrow$ 0.53) &37.01 ($\uparrow$ 1.31)  &21.88 ($\uparrow$ 3.13) &25.00 ($\uparrow$ 1.56) \\
	Gated Fusion + ViT     &35.08 ($\uparrow$ 9.15) &42.48 ($\uparrow$ 8.06)  &25.46 ($\uparrow$ 2.89) &41.73 ($\uparrow$ 6.03)  &25.00 ($\uparrow$ 6.25) &31.25 ($\uparrow$ 7.81) \\
	Selective Attn + ViT   		&\B{51.20} ($\uparrow$ \B{25.27})	&\B{64.71} ($\uparrow$ \B{30.29}) &\B{31.76} ($\uparrow$ \B{9.19})	&\B{53.54} ($\uparrow$ \B{17.84}) 	&\B{43.75} ($\uparrow$ \B{25.00}) &\B{56.25} ($\uparrow$ \B{32.81})\\
	\hline
	\multicolumn{7}{c}{\textit{English$\to$French}} \\
	\hline
	Text-only Transformer  &30.72 					&33.12 					  &34.91 				   &38.85    				 &23.44  				  &29.69\\
	Gated Fusion + ResNet  &32.68 ($\uparrow$ 1.96) &35.51 ($\uparrow$ 2.39) &32.55 ($\downarrow$ 2.36) &35.17 ($\downarrow$ 3.68) &17.19 ($\downarrow$ 6.25) &23.44 ($\downarrow$ 6.25) \\
	Gated Fusion + ViT     &45.53 ($\uparrow$ 14.81) &50.76 ($\uparrow$ 17.64) &45.41 ($\uparrow$ 10.50) &52.23 ($\uparrow$ 13.38) &34.38 ($\uparrow$ 10.94) &43.75 ($\uparrow$ 14.06) \\
	Selective Attn + ViT   		&\B{62.96} ($\uparrow$ \B{32.24})	&\B{68.85} ($\uparrow$ \B{35.73}) &\B{49.34} ($\uparrow$ \B{14.43}) &\B{55.38} ($\uparrow$ \B{16.53})   &\B{43.75} ($\uparrow$ \B{20.31}) 	&\B{53.12} ($\uparrow$ \B{23.43})\\
	\hline
	\end{tabular}
	}
	\caption{The accuracy of MMT systems when applied color-based probing.}\label{tab:color_probing}
  \end{table*}

  \begin{table*}[t]
	\resizebox*{\linewidth}{!}{
	\centering
	\begin{tabular}{l|ll|ll|ll}
	\hline
	\multirow{2}{*}{\textbf{Systems}} & \multicolumn{2}{c|}{\textbf{Test2016}} & \multicolumn{2}{c|}{\textbf{Test2017}} & \multicolumn{2}{c}{\textbf{MSCOCO}} \\
	\cline{2-7}
	& \textbf{Restrict} & \textbf{Relaxed} & \textbf{Restrict} & \textbf{Relaxed} & \textbf{Restrict} & \textbf{Relaxed}\\
	\hline
	\multicolumn{7}{c}{\textit{English$\to$German}} \\
	\hline
	Text-only Transformer  &59.49 					&64.05 					&58.56 						&62.53 						&60.94 					&65.62\\
	Gated Fusion + ResNet  &60.06 ($\uparrow$ 0.57) &64.91 ($\uparrow$ 0.86) &56.08 ($\downarrow$ 2.48) &59.06 ($\downarrow$ 3.47) &61.72 ($\uparrow$ 0.78) &65.23 ($\downarrow$ 0.39) \\
	Gated Fusion + ViT     &66.33 ($\uparrow$ 6.84) &70.76 ($\uparrow$ 6.71) &67.00 ($\uparrow$ 8.44) &71.46 ($\uparrow$ 8.93) 	 &71.09 ($\uparrow$ 10.15) &75.78 ($\uparrow$ 10.16) \\
	Selective Attn + ViT   &\B{73.04} ($\uparrow$ \B{13.55})&\B{78.89} ($\uparrow$ \B{14.84}) &\B{70.97} ($\uparrow$ \B{12.41})&\B{77.17} ($\uparrow$ \B{14.64}) &\B{73.44} ($\uparrow$ \B{12.50}) &\B{77.73} ($\uparrow$ \B{12.11})\\
	\hline
	\multicolumn{7}{c}{\textit{English$\to$French}} \\
	\hline
	Text-only Transformer  & 63.48						& 65.48 						& 61.04							& 62.53					 & 64.84							& 67.19\\
	Gated Fusion + ResNet  & 61.63 ($\downarrow$ 1.85)& 63.62 ($\downarrow$ 1.86) & 63.52 ($\uparrow$ 2.48)& 65.01 ($\uparrow$ 2.48) & 64.45 ($\downarrow$ 0.39)& 66.80 ($\downarrow$ 0.39)\\
	Gated Fusion + ViT     & 73.47 ($\uparrow$ 9.99)& 75.89 ($\uparrow$ 10.41) & 76.43 ($\uparrow$ 15.39)& 77.92  ($\uparrow$ 15.39)& \B{80.47} ($\uparrow$ \B{15.63})& \B{82.81} ($\uparrow$ \B{15.62})\\
	Selective Attn + ViT   &\B{78.89} ($\uparrow$ \B{15.41})&\B{81.31} ($\uparrow$ \B{15.83}) &\B{78.16} ($\uparrow$ \B{17.12})&\B{79.65} ($\uparrow$ \B{17.12}) & 79.69 ($\uparrow$ 14.85) & 81.64 ($\uparrow$ 14.45)\\
	\hline
	\end{tabular}
	}
	\caption{The accuracy of MMT systems when applied character-based probing.}\label{tab:character_probing}
  \end{table*}

Our implementation was based on Fairseq \cite{ott-etal-2019-fairseq}. For training, we used Adam Optimizer \cite{kingma2014adam} with $\beta_1=0.9$, $\beta_2=0.98$ and $\epsilon=10^{-8}$. We adopted the same learning rate schedule as \cite{vaswani2017attention}, where the learning rate first increased linearly for $warmup=2000$ steps from $1e^{-7}$ to $5e^{-3}$. After the warmup, the learning rate decayed proportionally to the inverse square root of the current step. Each training batch contained $4,096$ tokens. We also adopted the early-stop training strategy \cite{zhang2020UVR} to avoid the overfitting issue.

For evaluation, we averaged the last $10$ checkpoints for more reliable results. The width of beam size was set to $5$. The performance was measured by BLEU and METEOR for all test sets. Also, we used accuracy for evaluation on the probing tasks.

\subsection{Results}
Table \ref{tab:multi30k} summarizes the results on standard MMT data. Each model was evaluated on three test sets on two language pairs. We see, first of all, that the improvements of previous methods (Rows 2-4) over the tiny baseline are marginal in terms of both BLEU and METEOR. This confirms the assumption that the visual features are not fully used if the text is complete \cite{caglayan-etal-2019-probing}. When switching the vision features from ResNet (Row.5) to ViT (Row.6), there are no significant BLEU gains. Then, we test them on the proposed probing tasks to examine the ``real'' contribution to MMT.

\paragraph{Color-based Probing} Table \ref{tab:color_probing} shows the accuracy on the color-based probing task. We see that the accuracy improvement of the gated fusion method is marginal by both restrict and relaxed criteria. However, replacing ResNet with ViT yields gains of over $8$ accuracy points across three test sets on En-De task. Similar improvements are observed on the En-Fr task. The finding here indicates that stronger vision features are helpful for representing the visual information. Moreover, selective attention can make better use of the ViT features, achieving over $20$ accuracy gains on three test sets. This verifies the conjecture that the selective attention can further enhance the fused representation for the ViT features.

\input{fig_masking_main}

\begin{figure*}[!t]
	\centering
	\begin{tikzpicture}
  \node[draw,fill=cyan!10,minimum width=0.8em,minimum height=0.6em,label=left:\scriptsize{ViT\_Tiny:}] (c1) at (-4em,4.2em){};
  \node[draw,fill=cyan!30,minimum width=0.8em,minimum height=0.6em,anchor=north,label=left:\scriptsize{ViT\_Small:}] (c2) at ([yshift=-0.5em]c1.south){};
  \node[draw,fill=cyan!60,minimum width=0.8em,minimum height=0.6em,anchor=north,label=left:\scriptsize{ViT\_Base:}] (c3) at ([yshift=-0.5em]c2.south){};
  \node[draw,fill=cyan!90,minimum width=0.8em,minimum height=0.6em,anchor=north,label=left:\scriptsize{ViT\_Large:}] (c4) at ([yshift=-0.5em]c3.south){};

  \node[draw,fill=orange!10,minimum width=0.8em,minimum height=0.6em,label=left:\scriptsize{Swin\_Tiny:},anchor=east] (d1) at ([xshift=-4em]c1.west){};
  \node[draw,fill=orange!30,minimum width=0.8em,minimum height=0.6em,anchor=north,label=left:\scriptsize{Swin\_Small:}] (d2) at ([yshift=-0.5em]d1.south){};
  \node[draw,fill=orange!60,minimum width=0.8em,minimum height=0.6em,anchor=north,label=left:\scriptsize{Swin\_Base:}] (d3) at ([yshift=-0.5em]d2.south){};
  \node[draw,fill=orange!90,minimum width=0.8em,minimum height=0.6em,anchor=north,label=left:\scriptsize{Swin\_Large:}] (d4) at ([yshift=-0.5em]d3.south){};

  \scriptsize{
	\begin{axis}[
	at={(0,0)},
	  ybar,
	  height=.21\textwidth,
	  width=.21\textwidth,
	  bar width=1em,
	  tick align=inside,
	  ylabel style={yshift=-1em},
	  ylabel={BLEU},
	  symbolic x coords={ {mask-1}},
	  xtick=data,
	  ymin=34,ymax=37,
	  yticklabel style={/pgf/number format/precision=1,/pgf/number format/fixed zerofill}]
	  \addplot[fill=orange!10, draw,xshift=2.64em] coordinates {({mask-1},35.37)};
	  \addplot[fill=cyan!30,draw,xshift=2.64em] coordinates {({mask-1},36.52)};
	  \addplot[fill=cyan!60, draw,xshift=2.64em] coordinates {({mask-1},36.59)};
	  \addplot[fill=cyan!90, draw,xshift=2.64em] coordinates {({mask-1},36.0)};

 	  \addplot[fill=cyan!10,draw,xshift=-2.5em] coordinates {({mask-1},35.09)};
	  \addplot[fill=orange!30,draw,xshift=-2.5em] coordinates {({mask-1},35.45)};
	  \addplot[fill=orange!60, draw,xshift=-2.5em] coordinates {({mask-1},36.12)};
	  \addplot[fill=orange!90, draw,xshift=-2.5em] coordinates {({mask-1},35.78)};
	\end{axis}
  }

  \scriptsize{
	\begin{axis}[
	at={(11em,0)},
	  ybar,
	  height=.21\textwidth,
	  width=.21\textwidth,
	  bar width=1em,
	  tick align=inside,
	  symbolic x coords={{mask-2}},
	  xtick=data,
	  ymin=29,ymax=33,
	  yticklabel style={/pgf/number format/precision=1,/pgf/number format/fixed zerofill}]
	  \addplot[fill=cyan!10, draw,xshift=2.64em] coordinates { ({mask-2},31.82)};	
	  \addplot[fill=cyan!30, draw,xshift=2.64em] coordinates {({mask-2},31.96) };	
	  \addplot[fill=cyan!60, draw,xshift=2.64em] coordinates { ({mask-2},32.08)};
	  \addplot[fill=cyan!90, draw,xshift=2.64em] coordinates { ({mask-2},32.15)};
	
	  \addplot[fill=orange!10,draw,xshift=-2.5em] coordinates { ({mask-2},30.1)};	
	  \addplot[fill=orange!30,draw,xshift=-2.5em] coordinates {({mask-2},30.22) };
	  \addplot[fill=orange!60,draw,xshift=-2.5em] coordinates { ({mask-2},30.91)};
	  \addplot[fill=orange!90,draw,xshift=-2.5em] coordinates { ({mask-2},30.9)};
	
	\end{axis}
  }

  \scriptsize{
	\begin{axis}[
	at={(22em,0)},
	  ybar,
	  height=.21\textwidth,
	  width=.21\textwidth,
	  bar width=1em,
	  tick align=inside,
	  symbolic x coords={{mask-3}},
	  xtick=data,
	  ymin=26,ymax=30,
	  yticklabel style={/pgf/number format/precision=1,/pgf/number format/fixed zerofill}]
	  \addplot[fill=cyan!10, draw,xshift=2.64em] coordinates { ({mask-3},28.28)};
	  \addplot[fill=cyan!30, draw,xshift=2.64em] coordinates {({mask-3},28.63)};
	  \addplot[fill=cyan!60, draw,xshift=2.64em] coordinates {({mask-3},29.47)};
	  \addplot[fill=cyan!90, draw,xshift=2.64em] coordinates {({mask-3},28.86)};
	
	  \addplot[fill=orange!10,draw,xshift=-2.5em] coordinates { ({mask-3},27.17)};
	  \addplot[fill=orange!30,draw,xshift=-2.5em] coordinates {({mask-3},26.9)};
	  \addplot[fill=orange!60,draw,xshift=-2.5em] coordinates {({mask-3},27.52)};
	  \addplot[fill=orange!90,draw,xshift=-2.5em] coordinates {({mask-3},27.22)};
	
	\end{axis}
  }

  \scriptsize{
	\begin{axis}[
	at={(33em,0)},
	  ybar,
	  height=.21\textwidth,
	  width=.21\textwidth,
	  bar width=1em,
	  tick align=inside,
	  symbolic x coords={{mask-4}},
	  xtick=data,
	  ymin=23,ymax=28,
	  yticklabel style={/pgf/number format/precision=1,/pgf/number format/fixed zerofill}]
	  \addplot[fill=cyan!10, draw,xshift=2.64em] coordinates {({mask-4},25.6)};
	  \addplot[fill=cyan!30, draw,xshift=2.64em] coordinates {({mask-4},26.9)};	
	  \addplot[fill=cyan!60, draw,xshift=2.64em] coordinates {({mask-4},27.29)};
	  \addplot[fill=cyan!90, draw,xshift=2.64em] coordinates {({mask-4},26.66)};
	
	  \addplot[fill=orange!10,draw,xshift=-2.5em] coordinates {({mask-4},23.87)};
	  \addplot[fill=orange!30,draw,xshift=-2.5em] coordinates {({mask-4},24.53)};
	  \addplot[fill=orange!60,draw,xshift=-2.5em] coordinates {({mask-4},25.89)};
	  \addplot[fill=orange!90,draw,xshift=-2.5em] coordinates {({mask-4},25.54)};
	\end{axis}
  }

	\end{tikzpicture}
	\caption{BLEU scores$ [\%]$ for MMT models with ViT/Swin in various capacities on En-De Test2016.}
	\label{fig:vit_capacities}
  \end{figure*}
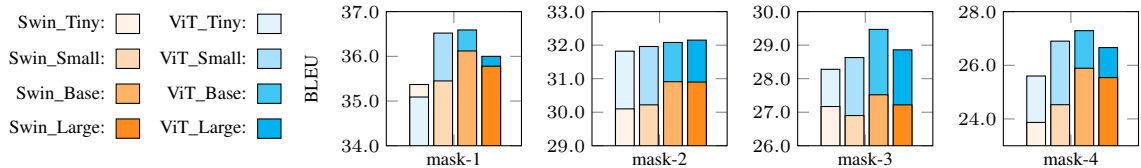

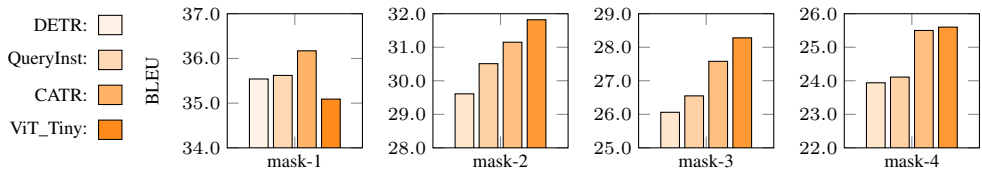
\begin{figure*}[!t]
	\centering
	\begin{tikzpicture}  
  \node[draw,fill=orange!10,minimum width=0.8em,minimum height=0.6em,label=left:\scriptsize{DETR:}] (d1) at (-4em,-2.8em){};
  \node[draw,fill=orange!30,minimum width=0.8em,minimum height=0.6em,anchor=north,label=left:\scriptsize{QueryInst:}] (d2) at ([yshift=-0.5em]d1.south){};
  \node[draw,fill=orange!60,minimum width=0.8em,minimum height=0.6em,anchor=north,label=left:\scriptsize{CATR:}] (d3) at ([yshift=-0.5em]d2.south){};
  \node[draw,fill=orange!90,minimum width=0.8em,minimum height=0.6em,anchor=north,label=left:\scriptsize{ViT\_Tiny:}] (d5) at ([yshift=-0.5em]d3.south){};

  \scriptsize{
	\begin{axis}[
	at={(0,-11em)},
	  ybar,
	  height=.21\textwidth,
	  width=.21\textwidth,
	  bar width=1em,
	  tick align=inside,
	  ylabel style={yshift=-1em},
	  ylabel={BLEU},
	  symbolic x coords={ {mask-1}},
	  xtick=data,
	  ymin=34,ymax=37,
	  yticklabel style={/pgf/number format/precision=1,/pgf/number format/fixed zerofill}]
	  \addplot[fill=orange!10, draw] coordinates {({mask-1},35.54)};   
	  \addplot[fill=orange!30,draw] coordinates {({mask-1},35.62)};
	  \addplot[fill=orange!60, draw] coordinates {({mask-1},36.17)};
	  \addplot[fill=orange!90, draw] coordinates {({mask-1},35.09)};
	\end{axis}
  }

  \scriptsize{
	\begin{axis}[
	at={(11em,-11em)},
	  ybar,
	  height=.21\textwidth,
	  width=.21\textwidth,
	  bar width=1em,
	  tick align=inside,
	  ylabel style={yshift=-2em},
	  symbolic x coords={{mask-2}},
	  xtick=data, 
	  ymin=28,ymax=32,
	  yticklabel style={/pgf/number format/precision=1,/pgf/number format/fixed zerofill}]
	  \addplot[fill=orange!20, draw] coordinates { ({mask-2},29.61)};
	  \addplot[fill=orange!35,draw] coordinates {({mask-2},30.51) };	  
	  \addplot[fill=orange!50, draw] coordinates { ({mask-2},31.15)};
	  \addplot[fill=orange!80, draw] coordinates {({mask-2},31.82)};
	 
	\end{axis}
  }
  
  \scriptsize{
	\begin{axis}[
	at={(22em,-11em)},
	  ybar,
	  height=.21\textwidth,
	  width=.21\textwidth,
	  bar width=1em,
	  tick align=inside,
	  ylabel style={yshift=-2em},
	  symbolic x coords={{mask-3}},
	  xtick=data,
	  ymin=25,ymax=29,
	  yticklabel style={/pgf/number format/precision=1,/pgf/number format/fixed zerofill}]
	  \addplot[fill=orange!20, draw] coordinates { ({mask-3},26.06)};	  
	  \addplot[fill=orange!35,draw] coordinates {({mask-3},26.55)};	  
	  \addplot[fill=orange!50, draw] coordinates {({mask-3},27.58)};
	  \addplot[fill=orange!80, draw] coordinates {({mask-3},28.28)};
	  
	\end{axis}
  }
  
  \scriptsize{
	\begin{axis}[
	at={(33em,-11em)},
	  ybar,
	  height=.21\textwidth,
	  width=.21\textwidth,
	  bar width=1em,
	  tick align=inside,
	  ylabel style={yshift=-2em},
	  symbolic x coords={{mask-4}},
	  xtick=data,
	  ymin=22,ymax=26,
	  yticklabel style={/pgf/number format/precision=1,/pgf/number format/fixed zerofill}]
	  \addplot[fill=orange!20, draw] coordinates {({mask-4},23.94)};
	  \addplot[fill=orange!35,draw] coordinates {({mask-4},24.11)};
	  \addplot[fill=orange!50, draw] coordinates {({mask-4},25.5)};
	  \addplot[fill=orange!80, draw] coordinates {({mask-4},25.6)};
	\end{axis}
  }
	\end{tikzpicture}
	\caption{BLEU scores $[\%]$ of various vision features on En-De Test2016.}
\label{fig:vision_objectives}
  \end{figure*}

\paragraph{Character-based Probing} Table \ref{tab:character_probing} shows similar results as in Table \ref{tab:color_probing}. ViT with selective attention performs the best on most scenarios, it is only slightly inferior to Gated Fusion + ViT on the MSCOCO dataset. While the gated fusion method with ResNet feature behaves far from desirable. It even underperforms the text-only Transformer, though the text-only Transformer is carefully regularized. A potential explanation is the character-based probing task is more challenging than the color-based probing task because it is more difficult for the model to find the correct corresponding region of the masked character word and provide useful signals to the text encoder.

\paragraph{Noun-based Probing}
Figure \ref{fig:comparison_main} plots the results of noun-based masking. It again verifies the above conjecture. The histograms in blue and red denote the results on the En-De and En-Fr tasks, respectively. The ViT features can significantly outperform the ResNet features across all masking methods on the two language pairs. We also observe that the gap between the ResNet and ViT features is gradually enlarged as more nouns are masked. This confirms the results in \cite{dosovitskiy2021ViT}.

\section{Analysis}

\subsection{How Vision Features Improve the MMT}
We further explore the impact of model capacity. Here, we report the results of ViT and Swin Transformer because they are strong models in recent studies. Our conjecture here is that larger ViT/Swin models can describe the image more accurately, which enables the text encoder to receive richer complementary information. Figure \ref{fig:vit_capacities} depicts the BLEU scores in progressive noun masking scenarios. Intuitively, larger ViT and Swin models provide more complementary knowledge to complete the insufficient text representations.

\begin{table*}[t]
	\resizebox*{\linewidth}{!}{
	\small
	\centering
	\begin{tabular}{l|c|c|r|cc|cc|cccc}
	\hline
	\multirow{2}{*}{\textbf{System}} & \multirow{2}{*}{\textbf{Patch}} &\multirow{2}{*}{\textbf{Reso. }} &\multirow{2}{*}{\textbf{Leng. }} & \multicolumn{2}{c|}{\textbf{Color Probing}} & \multicolumn{2}{c|}{\textbf{Character Probing}}& \multicolumn{4}{c}{\textbf{Noun Probing}} \\
	\cline{5-12}
	& & & & \textbf{Restrict} & \textbf{Relaxed} & \textbf{Restrict} & \textbf{Relaxed} & \textbf{$\textrm{Mask}_1$}  & \textbf{$\textrm{Mask}_2$}  & \textbf{$\textrm{Mask}_3$}  & \textbf{$\textrm{Mask}_4$} \\
	\hline
    ViT   &16$\times$16 & 384  &576 &49.67 &64.49          &74.32 &79.46       &36.59 &32.08 &29.47 &27.29 \\
	ViT   &16$\times$16 & 224  &196 &50.11 &61.87          &68.47 &74.32       &36.27 &31.49 &29.70 &26.51 \\
	ViT   &32$\times$32 & 384  &144 &49.02 &63.18          &70.19 &76.03       &35.53 &30.50 &28.28 &26.20 \\
	ViT    &32$\times$32 & 224  &49  &48.80 &61.00          &68.19 &73.47       &35.14 &30.30 &28.12 &25.19 \\
	\hline
	Swin    &4$\times$4 & 224  &49  &43.57 &54.47          &70.04 &75.18       &36.12 &30.91 &27.52 &25.89 \\
	\hline
	\end{tabular}
	}
	\caption{Comparison of various resolutions and patch sizes on the En-De (Test2016) probing tasks.}\label{tab:ablation_image}
  \end{table*}

\input{fig_examples}

Nevertheless, a counterintuitive phenomenon is the inferiority of Swin across all scenarios in the same configuration, though it outperforms ViT on most computer vision benchmarks. We attribute the reason to the short length of the patch sequence. In patch, ViT has a length of 577 (576 sequence segments and a special token \texttt{CLS}) when the image resolution and the patch size are $384 \times 384$ and $16 \times 16$. However, Swin has a fixed sequence length (49) restricted by the shifted window operation. This leads to more fine-grained local features for ViT, which is beneficial to the selective attention mechanism for extracting more relevant pieces.

\subsection{Impact of Learning Objectives}
Then, we investigate the impact of the enhanced vision features on MMT. Previous studies have already attempted to leverage object-detection features \cite{zhao-etal-2020-double,wang2021efficient} but the observation here is slightly different. Beyond the object-detection pretrained features, we also take the image captioning task into account.

Rows 11-13 in Table \ref{tab:multi30k} summarize the results of the three enhanced vision features on the standard MMT data, and Figure \ref{fig:vision_objectives} depicts the results on insufficient texts. Here we choose ViT-Tiny-based models for comparison due to the similar model capacity they own\footnote{Only pretrained vision models in a 256 hidden-size are available}. We see that not only the object-detection (DETR and QueryInst), but also the image captioning (CATR) pretrained features obtain superior performance compared with ViT-tiny (Row 8) when the text is complete. It is consistent with previous findings \cite{yin-etal-2020-novel,zhao-etal-2020-double}. However, the advantages do not persist when switching to limited text scenarios. A possible explanation is that these methods are sensitive to the quality of the extracted objects. We leave this as future work.

\begin{table*}[t]
	\small
	\resizebox*{\linewidth}{!}{
	\centering
	\begin{tabular}{l|l|c|c|c|c|c|c|c|c}
	\hline
	\multicolumn{2}{c|}{\multirow{2}{*}{\textbf{System}}} & \multicolumn{2}{c|}{\textbf{$\textrm{Mask}_1$}} & \multicolumn{2}{c|}{\textbf{$\textrm{Mask}_2$}} & \multicolumn{2}{c|}{\textbf{$\textrm{Mask}_3$}} & \multicolumn{2}{c}{\textbf{$\textrm{Mask}_4$}} \\
	\cline{3-10}
	\multicolumn{2}{c|}{} & \bf Cong. & \bf Icong. & \bf Cong. & \bf Icong. & \bf Cong. & \bf Icong. & \bf Cong. & \bf Icong.\\
	\hline
	\multicolumn{2}{c|}{Transformer-Tiny}  &34.37 &- &29.12 &- &24.03 &- &21.64 &-\\
	\hline
	Gated Fusion + ResNet  &  Pretrained &34.90 &34.88 	&28.94 &28.08  	&24.18 &22.56	&21.74  &20.79\\
	\hline
	Gated Fusion + ViT &  Pretrained    &35.61  &33.77 &30.40   &25.43  &27.58  &19.79 &25.30  &16.66\\
	\hline
	\multirow{2}{*}{Selective Attn + ViT}
	& \multicolumn{1}{l|}{Pretrained}   &36.59 &32.88	&32.08 &25.58	&29.47 &20.42	&27.29 &15.80\\
	& \multicolumn{1}{l|}{Scratch}      &34.91 &34.81	&28.91 &28.91	&23.40 &23.40	&19.63 &19.63\\
	\hline
	Selective Attn + DETR &  Pretrained &35.54  &33.92 &29.61   &27.20  &26.06  &21.65 &23.94  &18.88\\
	\hline
	Selective Attn + CATR &  Pretrained &36.17  &33.13 &31.15   &26.40  &27.58  &20.72 &25.50  &16.98\\
	\hline
	Select. Attn + ViT + CATR & Pretrained  	&\B{36.97} &32.98	&\B{32.45} &24.71	&\B{30.30} &19.92	&\B{28.14} &16.09\\
	\hline
	\end{tabular}
	}
	\caption{The impact of incongruent decoding for the noun masking strategy. Here Cong./Icong. denotes congruent and incongruent decoding, respectively. The results (BLEU $[\%]$) were measured on En-De Test2016.}\label{tab:incongruence}
  \end{table*}

\subsection{Impact of Resolution and Patch Size}
It is well-known that higher resolutions are beneficial to the accuracy improvement in computer vision tasks \cite{dosovitskiy2021ViT}. Despite the success of the Transformer architecture, recent studies show that the success of ViT mainly comes from the successful use of the patch schema \cite{dosovitskiy2021ViT}. Here, we compare MMT systems with different resolutions and patch sizes based on ViT-Base. The results on three probing tasks (see Table \ref{tab:ablation_image}) again confirm the above assumption that fine-grained vision features are more suited for the selective attention. Also, the attention map visualized in Figure \ref{fig:examples} demonstrates that high resolution with fine-grained patch schema can attend to correct regions of the image for each masked token. For example, both models pay the right attention to the masked character and noun, but the model with low resolution fails to detect the right region of color. The finding here may shed light to other multimodal tasks, such as VQA.

\begin{table*}[htbp!]
	\renewcommand{\arraystretch}{1.0}
	\centering
	\resizebox{.95\textwidth}{!}{%
	\begin{tabular}{cl@{}}
	\toprule
	\MR{7}{*}{\includegraphics[height=3.0cm]{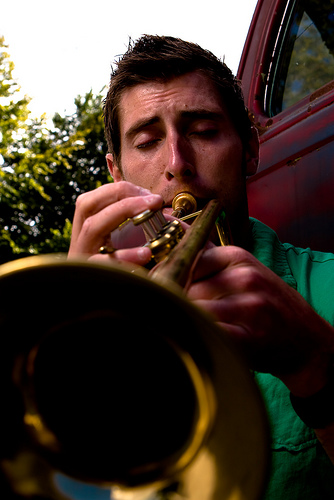}} &
	\T{SRC:} a brown-haired [man] in a [green] [shirt] plays a [trumpet] outdoors . \\
	& \T{REF:} ein \B{mann} mit braunen haaren in einem {\color{ForestGreen}{\B{grünen}}} \B{hemd} spielt im freien \B{trompete} . \\
	& \T{MK :} a brown-haired [$\texttt{MASK\_P}$] in a {\color{ForestGreen}{[$\texttt{MASK\_C}$]}} [$\texttt{MASK\_N}$] plays a [$\texttt{MASK\_N}$] outdoors . \\
	& \T{CNN:} eine braunhaarige \st{frau} in einem {\color{OrangeRed}{\st{roten}}} \st{kleid} spielt im freien \st{gitarre} . \\
	& \phantom{\T{CNN:}} \I{(a brown-haired \st{woman} in a {\color{OrangeRed}{\st{red}}} \st{dress} plays a \st{guitar} outdoors.)} \\
	& \T{ViT:} ein braunhaariger \B{mann} in einem {\color{ForestGreen}{\B{grünen}}} \B{hemd} spielt im freien \B{trompete} .\\
	& \phantom{\T{ViT:}} \I{(a brown-haired \B{man} in a {\color{ForestGreen}{\B{green}}} \B{shirt} plays a \B{trumpet} outdoors.)} \\

	\midrule
	\MR{7}{*}{\includegraphics[height=2.5cm]{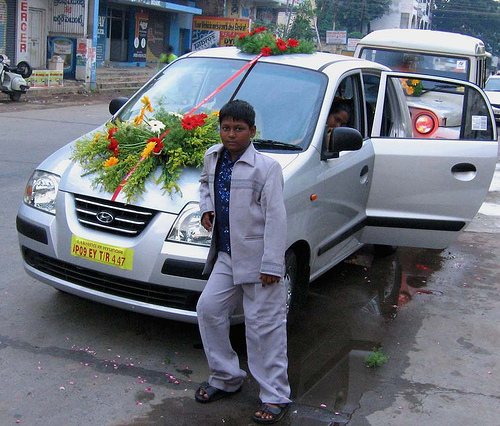}} &
	\T{SRC:} a [boy] is leaning on a [car] with [flowers] on the [hood] . \\
	& \T{REF:} ein junge lehnt sich an ein auto mit blumen auf der motorhaube . \\
	& \T{MK :} a [$\texttt{MASK\_P}$] is leaning on a [$\texttt{MASK\_N}$] with [$\texttt{MASK\_NS}$] on the [$\texttt{MASK\_N}$] . \\
	& \T{CNN:} ein \st{mann} lehnt an einer \st{wand} mit \sout{bäumen} auf der \sout{straße} . \\
	& \phantom{\T{CNN:}} \I{(a \st{man} is leaning on a \st{wall} with \st{trees} on the \st{street}.)} \\
	& \T{ViT:} ein \underline{kind} lehnt sich an einem \B{auto} mit \B{blumen} auf dem \st{gehweg} .\\
	& \phantom{\T{ViT:}} \I{(a \underline{child} is leaning on a \B{car} with \B{flowers} on the \st{sidewalk}.)} \\
	
	\bottomrule
	\end{tabular}}
	\caption{Qualitative examples from two complex scenarios. \st{Strikethrough} and \B{bold} words present the incorrect and good lexical choices. \underline{Underline} denotes the acceptable but not totally right translation.}
	\label{tab:case_study}
	\end{table*}
	
\subsection{Incongruent Decoding}

Incongruent decoding is a widely used manner to evaluate whether the visual modality contributes to the text \cite{caglayan-etal-2019-probing,caglayan-etal-2021-cross}. Table \ref{tab:incongruence} shows that incongruent decoding causes obvious BLEU drops except for the ResNet feature.
ViT beats the ResNet with gated fusion. It yields higher BLEU scores with congruent decoding and exhibits a larger BLEU drop with incongruent decoding.
We also find that the ViT features learned from scratch are also insensitive to the visual modality. This is reasonable that the learned vision systems are not sufficiently strong due to the data scarcity of Multi30K. Thus the visual modality acts more like noise signals. In addition, focusing on the results of pretrained selective attention + ViT, the gap between congruent and incongruent decoding gradually becomes larger.

We also investigate whether the ensemble vision features can help. Concretely, we choose ViT and CATR to independently generate the fused representations with the text feature, and then the ensemble feature is obtained based on them. We see that the ensemble vision feature performs the best on the congruent decoding, and achieves the largest BLEU gaps on four masking scenarios compared with other systems. These results again indicate that stronger visual contexts indeed help.

\subsection{Case Study}
Finally, we compare several real cases. We choose gated fusion (\T{CNN}) \cite{wu-etal-2021-good} and selective attention + ViT\_Base (\T{ViT}) for comparison. The qualitative examples in Table \ref{tab:case_study} demonstrate that the visual modality is complementary rather than redundant if the text is insufficient. To figure out whether the German translation is right or not, we provide the human-translation results. First, we see the top half case of Table \ref{tab:case_study}, \T{ViT} can fill in the masked entities and generate the correct translations even four entities were masked. Unfortunately, \T{CNN} incorrectly judges the man as a woman. Also, it cannot distinguish the right color of shirt due to the complex background. When given a more complex image (the bottom half case), it is still a challenge for ViT to generate the right translation. The observation here inspires us to design a more powerful fusion method. Also, the data scarcity problem is a root issue to prevent us from further improving the cross-modal translation quality.

\section{Related Work}
Multimodal machine translation is a cross-domain task in the field of machine translation. Early attempts mainly focused on enhancing the MMT model by better incorporation of the vision features \cite{calixto-liu-2017-incorporating,elliott-kadar-2017-imagination,delbrouck-dupont-2017-empirical}. However, directly encoding the whole image feature brings additional noise to the text \cite{yao-wan-2020-multimodal,liu2021gumbel}. To address the above issue, \citet{yao-wan-2020-multimodal} proposed a multimodal self-attention to consider the relative difference of information between two modalities. Similarly, \citet{liu2021gumbel} used a Gumbel Softmax to achieve the same goal.

Researchers also realize that the visual modality may be redundant. Irrelevant images have little impact on the translation quality, and no significant BLEU drop is observed even the image is absent \cite{elliott-2018-adversarial}.  Encouraging results appeared in \citet{caglayan-etal-2019-probing}'s work. They pointed out that the visual modality is still useful when the linguistic context is scarce, but is less sensitive when exposed to complete sentences. More recently, \citet{wu-etal-2021-good} attributed the BLEU gain on MMT tasks to the regularization training, and they again emphasized the imperative of constructing proper insufficient textual input. It is worthy to note that the proposed probing task is an improved version based upon previous work \cite{caglayan-etal-2019-probing,wu-etal-2021-good}. We also opensource the preprocessed data and the corresponding scripts for the subsequent researchers to experiment on.

Another line of research is to explore large-scale cross-modal pretraining models. In this way, the MMT task is regarded as a downstream task. For example, CLIP \cite{radford21clip} is a general cross-modal pretraining model, which learns to perform a wide variety of tasks via natural language prompting. \citet{caglayan-etal-2021-cross} presented a MMT-specific pretraining model which combines the translation language modeling with masked region classification objectives. In this work, we make a systematic study on whether stronger vision features are helpful. We also extend the research to enhanced features, such as object-detection and image captioning, which are complementary to previous work.

\section{Conclusions}
In this work, we show that stronger vision features (e.g. ViT-like models) strengthen MMT systems on three proposed probing tasks. We present a selective attention method for ViT-based models to make better use of the patch-level representation. The result here shows a promising line of research on developing better vision models for multimodal tasks. As far as we know, this is the first attempt to build MMT systems with Transformer only. In future work, we are willing to investigate whether it is possible to use a single set of parameters to encode the vision and text modalities.

\section*{Acknowledgments}
This work was supported in part by the National Science Foundation of China (Nos. 61732005 and 61876035), the National Key R\&D Project of China (No. 2019QY1801), the China HTRD Center Project (No. 2020AAA0107904) and Yunnan Provincial Major Science and Technology Special Plan Projects (Nos. 201902D08001905 and 202103AA080015). The authors would like to thank anonymous reviewers for their valuable comments. And thank Yufan Jiang for his helpful advice to improve the paper.

\bibliography{multimodal}
\bibliographystyle{acl_natbib}

\end{document}